\documentclass[a4paper,11pt]{article}
\usepackage{latexsym}
\usepackage{amsmath}
\usepackage{amssymb}
\usepackage{verbatim}
\usepackage{natbib}
\usepackage{multirow}
\usepackage{color}

\usepackage[utf8]{inputenc} 
\usepackage[T1]{fontenc}    
\usepackage{hyperref}       
\usepackage{url}            
\usepackage{booktabs}       
\usepackage{microtype}      
\usepackage{bm}

\newif\ifpdf
\ifx\pdfoutput\undefined
  \pdffalse                     
  \usepackage[dvips]{graphicx}
\else
  \pdfoutput=1                  
  \pdftrue
  \usepackage[pdftex]{graphicx}
\fi

\newcommand{\cut}[1]{}
\newcommand{\bfv}{\vec{v}}
\newcommand{\bfx}{\vec{x}}
\newcommand{\bfz}{\vec{z}}
\newcommand{\defeq}{{\stackrel{def}{=}}}

\renewcommand{\vec}[1]{\mathbf{#1}}

\title{Autoencoders and Probabilistic Inference with Missing Data: \\ 
An Exact Solution for The Factor Analysis Case}

\author{
  Christopher K.~I.~Williams \\
  School of Informatics \\
  University of Edinburgh, UK \\
  Alan Turing Institute, London, UK \\
  \texttt{c.k.i.williams@ed.ac.uk} \\
  \and
  Charlie Nash \\
  School of Informatics \\
  University of Edinburgh, UK \\
  \texttt{charlie.nash@ed.ac.uk} \\
  \and
  Alfredo Naz\'{a}bal \\
  Alan Turing Institute, 
  London, UK \\
  \texttt{anazabal@turing.ac.uk}
}

\date{\today}

\begin{document}

\maketitle

\begin{abstract}
  Latent variable models can be used to probabilistically ``fill-in''
missing data entries. The variational autoencoder architecture
\citep{kingma-welling-14,rezende-mohamed-wierstra-14} 
includes a ``recognition'' or ``encoder'' network that infers the latent variables
given the data variables. However, it is not clear how to handle
missing data variables in this network. 
The factor analysis (FA) model is a basic autoencoder, using linear
encoder and decoder networks.  We show how to
calculate exactly the latent posterior distribution 
for the FA model in the presence of missing data,
and note that this solution implies that a different encoder network
is required for each
pattern of missingness. We also discuss various approximations
to the exact solution. Experiments compare the effectiveness 
of various approaches to imputing the missing data.
\end{abstract}


\maketitle

\section{Introduction}\label{sec:introduction}

Latent variable models, like factor analysis and ``deeper'' versions
such as the variational autoencoder (VAE,
\citealt*{kingma-welling-14,rezende-mohamed-wierstra-14}) and generative
adversarial networks (GANs, \citealt*{goodfellow-etal-14}),
are a compelling approach to
modelling structure in complex high-dimensional data such as images.

One important use case of such models is when some part of the
observable data is missing---such as when some part of an image is not
observed. In this case we would like to use the latent variable model
to ``inpaint'' the missing data.  Another example is modelling 3-D
volumetric data---we may have observations of an object
from one viewpoint, and wish to make inferences about the whole 3-D
object.

The VAE is composed of two parts, an \emph{encoder} (or recognition)
network that predicts the distribution of the latent variables given
the data, and a \emph{decoder} (or generative) model that maps from the
latent variables to the visible variables. However, if some part of
the visible data is missing, how can we make use of the encoder
network---what can be done about the missing entries? One simple
fix is to replace the missing value with a constant such as zero
(see e.g.\ \citealt*{pathak-krahenbuhl-donahue-darrell-16},
\citealt*{nazabal-olmos-ghahramani-valera18}) or 
``mean imputation'', i.e.\ filling in the missing value
with the unconditional mean of that variable. However, these
are not principled solutions, and do not maintain uncertainty about
the missing values.

\citet{rezende-mohamed-wierstra-14} have shown that one
can construct a Markov chain Monte Carlo (MCMC) method to sample from
the posterior over the missing data for a VAE, but this is an
iterative technique that has the obvious disadvantage of taking a long
time (and much computation) for the chain to converge.

This paper is concerned with inference for the latent variables
given a particular pre-trained generative model. The issue of 
learning a model in the presence of missing data is a different
issue; for PCA a survey of methods is given e.g.\ in
\cite{ilin-raiko-10,dray-josse-15}.

The non-linear encoder and decoder networks of the VAE make an exact
analysis of missing-data inference techniques challenging. As such we
focus on the factor analysis (FA) model as an example autoencoder
(with linear encoder and decoder networks) for which analytic
inference techniques can be derived.  In this paper we show that for
the FA model one can carry out exact (non-iterative) inference for the
latent variables in a single feedforward pass. However, the parameters
of the required linear feedforward model depend non-trivially on the
missingness pattern of the data, requiring a different encoder
network for each pattern. Below we consider three approximations to
exact inference, including a ``denoising encoder'' approach to the
problem, inspired by the denoising autoencoder of
\cite{vincent-etal-08}. Experiments compare the effectiveness of
various approaches to filling in the missing data.

 

\section{Theory}
We focus on generative models which have latent
variables associated with each datapoint. We assume that such a model
has been trained on fully visible data, and that 
the task is to perform inference of the latent variables in the presence of
missing data, and to reconstruct the missing input data.

Consider a generative latent-variable model that consists of a prior
distribution $p(\vec{z})$ over latent variables $\vec{z}$ of dimension
$K$, and a conditional distribution over data variables given latents
$p(\vec{x} | \vec{z})$.
When there is missing data, the data
variables $\vec{x}$ can be separated into the visible variables
$\vec{x}_v$ and missing variables $\vec{x}_m$. 
$\vec{m}$ is a missing-data indicator function 
such that $m_j = 1$ if $x_j$ has been observed, and
$m_j = 0$ if $x_j$ is missing. 
Let $\vec{x}$ have dimension $D$, with $D_v$ visible and 
$D_m$ missing variables.

Given a datapoint $
[\vec{x}_v, \vec{m}]$ we would like to be able to obtain
a latent posterior conditioned on the visible variables $p(\vec{z} |
\vec{x}_v, \vec{m})$. We assume below that the missing data is 
\emph{missing at random} (MAR, see \citealt*{little-rubin-87}
for further details), i.e.\
that the missingness mechanism does not depend on the values
of the missing data.

One can carry out exact inference for linear subspace models such as
factor analysis and its special case probabilistic principal components
analysis (PPCA, \citealt*{tipping-bishop-99}). Let $p(\vec{z}) \sim
N(0,I_K)$ and $p(\vec{x}|\vec{z}) \sim N(W \vec{z} + \bm{\mu}, \Psi)$
where $W$ is the $D \times K$ factor loadings matrix, $\bm{\mu}$ is
the mean (offset) vector in the data space, and $\Psi$ is a
$D$-dimensional diagonal covariance matrix.
In these models the general form of the posterior is $p(\vec{z} |
\vec{x}_v, \vec{m}) = \mathcal{N}(\vec{z} | \bm{\mu}_{\vec{z} |
	\vec{x}_v}, {\Sigma}_{\vec{z} | \vec{x}_v})$, where
\begin{align}
{\Sigma}_{\vec{z} | \vec{x}_v} &= \left( {I}_K +
{W}^{\intercal}_{v} {\Psi}^{-1}_{v} {W}_{v}\right)^{-1}, \label{eq:covv}\\
\bm{\mu}_{\vec{z} | \vec{x}_v} &= {\Sigma}_{\vec{z} |
	\vec{x}_v} {W}^{\intercal}_{v} {\Psi}_{v}^{-1}(\vec{x}_v
- \bm{\mu}_v) , \label{eq:meanv}
\end{align}
see e.g.\ \cite[sec.\ 12.2]{bishop-06},
where $W_v$ denotes the submatrix of $W$ relating to the visible
variables,
and similarly for $\Psi_v$.
These equations are simply
the standard form for a FA model when the missing variables have been
marginalized out.
The expression for ${\Sigma}_{\vec{z} | \vec{x}_v}$ can be re-written as
\begin{equation}
{\Sigma}_{\vec{z} | \vec{x}_v} = (I_K + W^T M \Psi^{-1} M W)^{-1}  ,
\label{eq:covv2}
\end{equation}
where $M$ is a diagonal matrix the $j$th entry being $m_j$\footnote{Given
	that $M$ is idempotent i.e.\ $M^2 = M$, one does not need to make use of
	the usual $M^{1/2}$ construction on either side of $\Psi^{-1}$.}. This
means that the diagonal elements in $M \Psi^{-1} M$ will be
be $m_j \psi^{-1}_{jj}$: if $m_j = 1$ (so $x_j$ is visible) then
$x_j$ is observed with variance $\psi_{jj}$, 
but for missing data dimensions $m_j = 0$ implies
an effective infinite variance for $\psi_{jj}$, meaning that any data 
value for this missing dimension will be ignored.
Note also that an information-theoretic argument that
conditioning on additional variables never increases entropy shows that
${\Sigma}_{\vec{z} | \vec{x}}$ will have a determinant no larger than
${\Sigma}_{\vec{z} | \vec{x}_v}$\footnote{The entropy argument strictly applies when taking expectations over $\vec{x}$ or $\vec{x}_v$, but for Gaussian distributions the predictive uncertainty is independent of the
	value of $\vec{x}$ or $\vec{x}_v$, so it does imply the desired conclusion.}.

Equation \ref{eq:covv2} can be rewritten in an interesting
way.
Let $\bfv_j^{\intercal} = (w_{j1}, \ldots, w_{jK})$ denote the $j$th \emph{row}
of $W$. Then 
\begin{equation}
\Sigma^{-1}_{\vec{z} | \vec{x}_v} =
I_K + W^{\intercal} M {\Psi}^{-1} M W = I_K + \sum_{j=1}^D m_j \psi^{-1}_{jj}
\bfv_j  \bfv_j^{\intercal},  \label{eq:rank1sum}
\end{equation}
i.e.\ where the second term on the RHS is a sum of rank-1 matrices. This arises
from the fact that $p(\bfz|\bfx) \propto p(\bfz, \bfx) = p(\bfz)
\prod_{j=1}^D p(x_j|\bfz)$, and can be related to the product of
Gaussian experts construction \cite{williams-agakov-02}. Similarly
the mean $\bm{\mu}_{\vec{z} | \vec{x}_v}$ has the form
\begin{equation}
\bm{\mu}_{\vec{z} | \vec{x}_v} = \Sigma_{\bfz|\bfx_v} \sum_{j=1}^D
m_j \psi^{-1}_{jj} (x_j - \mu_j) \bfv_j  . \label{eq:rank1mean}
\end{equation}
Again note how each observed dimension contributes one non-zero term to the 
sum on the RHS. 

The non-identifiability of the FA model due to 
``rotation of factors'' (see e.g.\ 
\citealt*[sec.~9.6]{mardia-kent-bibby-79}) means one can transform $W$ with
an orthogonal matrix $U$ so that $p(\bfz|\bfx)$ is a diagonal Gaussian.
If $\Sigma_{\bfz|\bfx}= U \Lambda U^T$, then setting $\tilde{W} = 
W U$ makes the corresponding $\tilde{\Sigma} $ be diagonal.
\citet[sec.~2]{vedantam-fischer-huang-murphy-18} suggest that for a VAE with
missing data, one can combine together \emph{diagonal} Gaussians
in $\bfz$-space from each observed data dimension to obtain a posterior
$p(\bfz|\bfx_v)$ using a product of Gaussian experts construction. However,
our analysis above shows that this cannot be exact: even if we are in the
basis where $p(\bfz|\bfx)$ is a diagonal Gaussian, we note from eq.\
\ref{eq:rank1sum} that the contribution of each observed dimension is 
a rank-1 update to $\Sigma^{-1}_{\vec{z} | \vec{x}_v}$, and thus
there will in general be no basis in which all of these updates will
be diagonal. However, note that a rank-1 update involves the
same number of entries ($K$) as a diagonal update.

\begin{figure*}[t]
	\centering
	\includegraphics[width=6in]{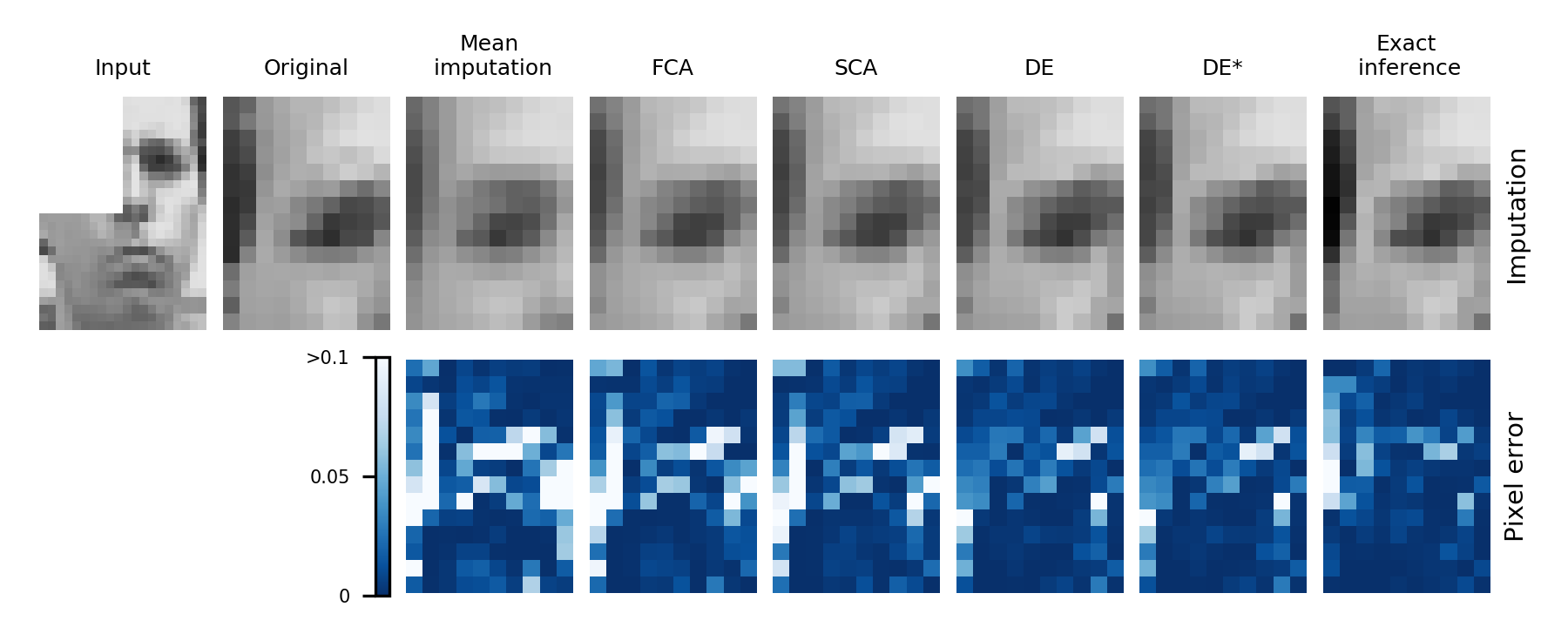}
	\caption{Illustration of the seven reconstruction methods on a 
		specific image (leftmost) with the top left quarter missing. Top
		row: second column shows the original quarter, columns 3-7 show the
		corresponding imputations. The bottom row shows the pixelwise squared error.
		\label{fig:freyex}}
\end{figure*}

\begin{figure*}[t]
	\centering
	\includegraphics[width=6.25in]{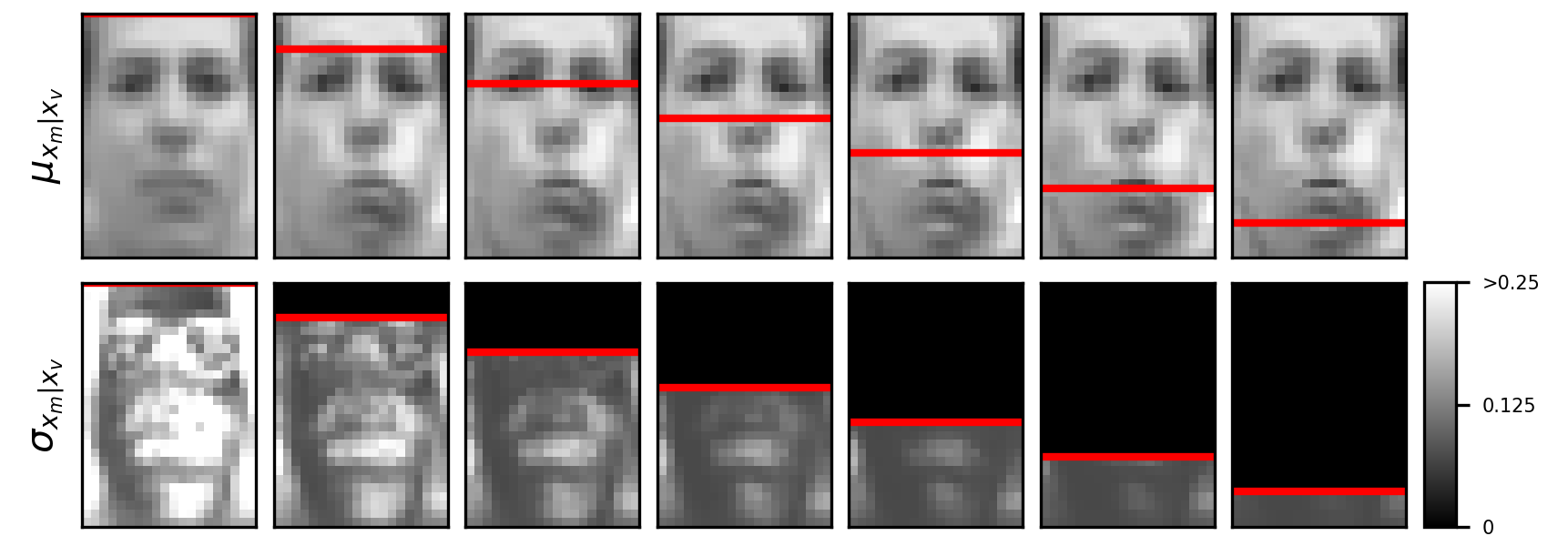}
	\caption{%
		Figure  illustrating how the posterior means and
		standard deviations
		for unobserved pixels change as the amount of missing data decreases. Data 
		above the red line is present, below it is missing and is imputed
		with exact inference.
		\label{fig:freyvar}}
\end{figure*}

If the $\vec{x}$ data does not contain missing values, it is
straightforward to build a ``recognition network'' that predicts
$\bm{\mu}_{\vec{z} | \vec{x}}$ and ${\Sigma}_{\vec{z} | \vec{x}}$
from $\vec{x}$ using only matrix-vector multiplies, as per equations
\ref{eq:covv} and \ref{eq:meanv} without the $v$
subscripts, as the matrix inverse can be computed once and stored.
However, if there is missing data then there is a
different ${\Sigma}_{\vec{z} | \vec{x}_v}$ for each of the $2^D
-1$ non-trivial patterns of missingness, and thus exact inference cannot be
carried out efficiently in a single feedforward inference network. 
This leads to a number of approximations:

\vspace*{2mm}

\paragraph{Approximation 1: Full Covariance Approximation (FCA)}
If we carry out mean imputation for $\vec{x}_m$ and replace those
entries with $\bm{\mu}_m$ to give $\vec{x}^{mi}$, then we can rewrite
eq.\ \ref{eq:meanv} as
\begin{equation}
\bm{\mu}_{\vec{z} | \vec{x}_v} = \Sigma_{\bfz|\bfx_v} 
{W}^{\intercal} {\Psi}^{-1}(\vec{x}^{mi} - \bm{\mu}), \label{eq:meanmi}
\end{equation}
as the missing data slots in $\vec{x}^{mi} -\bm{\mu}$ will be filled in by
zeros and have no effect on the calculation. However, note that 
${\Sigma}_{\vec{z} | \vec{x}_v}$ \emph{does} depend on the pattern of
missingness $\vec{m}$ as per eq.\ \ref{eq:covv2}. A simple approximation
is to replace it  with ${\Sigma}_{\vec{z} | \vec{x}}$ 
which would apply when $\vec{x}$ is fully observed; we term this the
\emph{full covariance approximation}, as it uses the covariance relating to
a \emph{fully observed} $\bfx$.
We denote this approximation as $\bm{\mu}^{FCA}_{\vec{z} | \vec{x}_v}$, as written in eq.\ \ref{eq:FCAapprox}:
\begin{equation}
\bm{\mu}^{FCA}_{\vec{z} | \vec{x}_v} = \Sigma_{\vec{z} | \vec{x}}
{W}^{\intercal} {\Psi}^{-1}(\vec{x}^{mi} - \bm{\mu}) . \label{eq:FCAapprox}
\end{equation}

This approximation uses a \emph{fixed} encoder network for all patterns of missingness.

\vspace*{2mm}
\paragraph{Approximation 2: Scaled Covariance Approximation (SCA)}
When no data is observed, we have that the posterior covariance
is equal to the prior, $I_K$. When there is complete data, then 
\begin{equation}
\Sigma^{-1}_{\bfz|\bfx} \defeq P_{\bfz|\bfx} = I_K + \sum_{j=1}^D P_j 
\label{eq:SCAapprox}
\end{equation}
where $P_j = \psi^{-1}_{jj} \bfv_j \bfv_j^{\intercal}$, making use of
eq.\ \ref{eq:rank1sum}.  
Although as we have seen the $P_j$'s are all different, a simple approximation
would be to assume they are all equal, and thus rearrange
eq.\ \ref{eq:SCAapprox} to obtain $\tilde{P}_j = (P_{\bfz|\bfx} -
I_K)/D$ for $j = 1, \ldots, D$.\footnote{A slightly more general
	decomposition would be to set $\tilde{P}_j = \alpha_j (P_{\bfz|\bfx} - I_K)$
	with $\sum_j \alpha_j = 1$, but then it would be more difficult to 
	decide how to set the $\alpha_j$s; this could e.g.\ be learned
	for a given missingness mechanism.} This would give the approximation
\begin{equation}
\Sigma^{-1}_{\bfz|\bfx_v} \simeq I_K + \frac{\sum_j m_j}{D}
(P_{\bfz|\bfx} - I_K) = \frac{D_m}{D} I_K + \frac{D_v}{D} P_{\bfz|\bfx}
, \label{eq:sca}
\end{equation}
which linearly interpolates between $I_K$ and $P_{\bfz|\bfx}$
depending on the amount of missing data. Using this approximation
instead of $\Sigma_{\bfz|\bfx}$ in eq.\ \ref{eq:FCAapprox} yields the
\emph{scaled covariance approximation} (SCA). The inversion of the RHS
of eq.\ \ref{eq:sca} can be carried out simply and analytically if
$P_{\bfz|\bfx}$ is diagonal, as per the rotation-of-factors discussion
above.
The SCA approximation does not use a fixed encoder network, but
the adjustment can be achieved very cheaply via multiplication by
a $K$-dimensional diagonal matrix depending on $D_m/D$.

\vspace*{3mm}

\paragraph{Approximation 3: Denoising Encoder (DE)}
\citet{vincent-etal-08} introduced the \emph{denoising
  autoencoder} (DAE), where the goal is to train an autoencoder to
take a corrupted data point as input, and predict the original,
uncorrupted data point as its output. In \cite{vincent-etal-08} the
stochastic corruption process sets a randomly-chosen set of the inputs
to zero; the goal is then to reconstruct the corrupted inputs from the
non-corrupted ones.  As we have seen above, if the data is centered,
then setting input values to zero is equivalent to mean
imputation. Despite this, the motivation for denoising autoencoders
was not so much to handle missing data, but to more robustly learn
about the structure of the underlying data manifold. Also, as seen
from our results above, a simple feedforward net that ignores the
pattern of missingness (as in the DAE) cannot perform exact inference
even for the factor analysis case.

However, we can develop ideas relating to the DAE to create a
\emph{denoising encoder} (DE). In our situation we have complete
training data (i.e.\ without missingness). Thus for each $\bfx$ 
we can obtain the corresponding posterior for $\bfz$ (as per 
eqs.\ \ref{eq:covv} and \ref{eq:meanv} without any missing data). We then 
create a pattern of missingness and apply it to $\bfx$ to obtain
$\bfx^{mi}$. One can then train a regression model from 
$\bfx^{mi}$ to the corresponding posterior mean of $\bfz$. Note that 
this \emph{averages} over the patterns of missingness, rather than
handling each one separately. The \emph{decoder} for the DE is the exact 
solution $p(\vec{x}|\vec{z}) \sim N(W \vec{z} + \bm{\mu}, \Psi)$.
The DE approximation uses a fixed encoder network for all
  patterns of missingness.  However, a limitation of the DE is that
to train it we need examples of the patterns of missingness that occur
in the data, which is not the case for the exact or approximate
methods described above.


Note: In February 2019 we became aware of the paper
by \citet{ilin-raiko-10}.  These authors present similar equations
to our eqs.\ \ref{eq:rank1sum} and \ref{eq:rank1mean} for the PPCA 
case in their eq.\ 17, and also discuss in their sec.\ 6.3 how PPCA
inference for complete data can be carried out with a diagonal
Gaussian. However, they did not take an autoencoder
view of their work, nor consider and evaluate the three
approximations described above.

\section{Experiments}
\newcommand{\colwidth}{22mm}
\begin{table*}
	\caption{Mean squared imputation error ($\times 10^{2}$) with
          standard errors for various imputation methods on PPCA
          generated data. For a particular data example the squared
          imputation error is the average error over imputed values
          for that example. Denoising encoder$^*$ is trained on random
          patterns of missingness, whereas Denoising encoder is
          trained on the same type of missingness patterns as it is
          tested on. For the Denoising encoder and Denoising encoder$^*$
 on the Random data, note that the performance is identical as the training and testing data are the same in each case. This is indicated with brackets. \label{tab:ppca-frey}}
	\centering
        \vspace*{2mm}
	\footnotesize
	\begin{tabular}{llp{\colwidth}p{\colwidth}p{\colwidth}p{\colwidth}p{\colwidth}p{\colwidth}}
		\toprule
		&        Mean \mbox{imputation} & Full-covariance \mbox{approx.} & Scaled-covariance \mbox{approx.} & Denoising \mbox{encoder} & Denoising \mbox{encoder*} & Exact \mbox{inference} \\ \midrule
		Random   &  $4.5323 \pm 0.0047$       & $1.8675 \pm 0.0017$   & $0.9783 \pm 0.0007$       & $1.0330 \pm 0.0006$      & ($1.0330 \pm 0.0006$)     & $0.5913 \pm 0.0001$          \\   
		Quarters &  $4.4642 \pm 0.0065$       & $2.5732 \pm 0.0045$  &  $2.2671 \pm 0.0041$     & $1.0043 \pm 0.0011$     &   $1.6863 \pm 0.0021$   & $0.6998 \pm 0.0005$         \\
		\bottomrule
	\end{tabular}
\end{table*}

\begin{table*}
	\caption{Mean squared imputation error ($\times 10^{2}$) with standard errors for various imputation methods on the real Frey faces dataset. The details are as in Table \ref{tab:ppca-frey}. \label{tab:frey}}
	\centering
	\footnotesize
	\label{my-label}
	\begin{tabular}{llp{\colwidth}p{\colwidth}p{\colwidth}p{\colwidth}p{\colwidth}p{\colwidth}}
		\toprule
		&       & Mean \mbox{imputation} & Full-covariance \mbox{approx.} & Scaled-covariance \mbox{approx.} & Denoising \mbox{encoder} & Denoising \mbox{encoder*} & Exact \mbox{inference} \\ \midrule
		\multirow{2}{*}{Random}   & Train & $4.6390 \pm 0.0013$       & $1.9339 \pm 0.0006$      &  $1.0433 \pm 0.0004$  & $0.7161 \pm 0.0002$      & ($0.7161 \pm 0.0002$)     & $0.6583 \pm 0.0002$          \\
		& Test  & $4.6218 \pm 0.0054$       & $1.9531 \pm 0.0026$    &  $1.0878 \pm 0.0018$   & $1.1450 \pm 0.0022$     & ($1.1450 \pm 0.0022$)      & $0.7165 \pm 0.0016$          \\   
		\rule{0pt}{4ex}\multirow{2}{*}{Quarters} & Train & $4.6002 \pm 0.0017$       & $2.7613 \pm 0.0012$    & $2.4820 \pm 0.0011$    & $0.7262 \pm 0.0003$     &   $1.8731 \pm 0.0009$   & $1.2507 \pm 0.0007$         \\ 
		& Test   & $4.4886 \pm 0.0066$       & $2.7124 \pm 0.0044$    &   $2.4496 \pm 0.0040$   & $1.2220 \pm 0.0036$      &   $1.8883 \pm 0.0036$   & $1.3677 \pm 0.0035$          \\
		\bottomrule
	\end{tabular}
\end{table*}
We trained a PPCA model on the Frey faces dataset\footnote{Available 
	from \url{https://cs.nyu.edu/~roweis/data/frey_rawface.mat}.},
which consists of 
1965 frames of a greyscale video sequence with resolution $20 \times 28$ 
pixels. Pixel intensities were rescaled to lie between -1 and 1. We used 
43 latent components for the model, explaining $90\%$ of the data 
variability. $80\%$ of the data frames were randomly selected as a 
training set, and the remainder were held out as a test set.

We estimate the parameters $W, \bm{\mu}, \sigma^2$
of the model using maximum likelihood, and obtain the analytic solutions
as in \cite{tipping-bishop-99}.
In the case of the denoising encoder, we create one
missingness-corrupted pattern $\bfx^{mi}$ for each training example 
when estimating the regression model.

Given a trained PPCA model we investigate various approaches for 
handling missing data with a data imputation task. Given a partially 
observed data example, the task is to predict the values of the missing 
data variables using the trained model. We consider two patterns of 
missing data. In the first setting each data variable is independently 
dropped out with probability $0.5$. In the second setting one of the four 
image quarters is dropped out in each example (see Fig.\ \ref{fig:freyex}).

We consider five approaches to handling missing data. The first method
is simply to inpaint the missing variables with their training set
means, without using the latent variable model at all. This is denoted
by ``mean imputation''. The second approach is to use FCA, and the
third SCA. For the fourth method we train a denoising encoder as
described above, using a linear regression model.  The final method is
to use exact inference as per eqs.\ \ref{eq:covv} and \ref{eq:meanv}.

\subsection{Results}
We present two sets of results, firstly for data generated from the
PPCA model fitted to the Frey data (denoted PPCA-Frey data), and
secondly for the Frey data itself.

For the {\bf PPCA-Frey data}, the mean squared imputation error
results are shown in Table \ref{tab:ppca-frey}. In this case 400 test
cases are generated from the PPCA model. For the DE case, 1600
training cases are also generated to train the DE model.  For the
denoising autoencoder on the quarters data, we train it
either on data with missing quarters, or on data with pixels with a
random pattern of missingness---the latter is denoted as Denoising
encoder$^*$ in the tables.

In Table \ref{tab:ppca-frey} we see that FCA improves over mean
imputation, and that SCA improves over FCA. Training the denoising
autoencoder (which uses the same input $\bfx^{mi}$ as FCA and SCA)
gives a further improvement for the quarters data, but is very similar
to SCA for the random missingness pattern.  Note that on the
quarters data, the Denoising encoder$^*$ results are quite a lot
worse than the DE which had been trained on missing-quarters training
data.  For both the random and quarters data, the exact inference
method is the best, as expected.

For the {\bf Frey data} the mean-squared imputation error results are
shown in Table \ref{tab:frey}, and a comparison on a specific image is
shown in Figure \ref{fig:freyex}. For the random missingness data,
mean imputation is the worst and exact inference the best, with FCA,
DE and SCA falling in between (from worst to best). Exact inference
significantly outperforms the denoising encoder on the test set.  For
the quarters dataset, we see a similar pattern of performance, except
for slightly better performance of the denoising encoder than exact
inference on the test set. Notice also that when using the Denoising
encoder$^*$ (training using random patterns of missingness) the
performance is significantly worse. The slight performance gain of the
DE on the quarters data over exact inference may be due to the fact
that exact inference applies to the PPCA model, but as the Frey data
was not generated from this model there can be some room for
improvement over the ``exact'' method.  Also note that for the
quarters data only four patterns of missingness are used, so the DE
network is likely to be able to learn about this more easily than
random missingness.
There are noticeable differences between the training and test set
errors for the DE model. This can be at least partially explained by
noting that on the training set the DE model has access to the
posterior for $\bfz$ based on the \emph{complete} data, so it is not
surprising that it does better here.

Figure \ref{fig:freyvar} illustrates how the 
posterior means and standard deviations of unobserved pixels change 
as the amount of missing data decreases, using exact inference. As
would be expected the uncertainty decreases as the amount of visible data 
increases, but notice how some regions like the mouth retain higher 
uncertainty until observed.

\section{Conclusions} 
Above we have discussed various approaches
to handling missing data with probabilistic autoencoders. We have shown
how to calculate exactly the latent posterior distribution for the
factor analysis (FA) model in the presence of missing data. This
solution exhibits a non-trivial dependence on the pattern of
missingness, requiring a different matrix inversion for each pattern.
We have also described three approximate inference methods (FCA, SCA and DE).
Our experimental results show the relative
effectiveness of these methods on the Frey faces dataset with
random or structured missingness. A limitation of the DE method
is that the structure of the missingness needs to be known at 
training time---our results demonstrate that performance 
deteriorates markedly when the missingness assumptions at train 
and test time differ. 

A possible future direction is to investigate whether these insights
can be used to inform modeling choices for nonlinear auto-encoders
trained in the presence of missing data. For instance, a variational
Gaussian posterior could use the structure as given in eqs.\
\ref{eq:rank1sum} and \ref{eq:rank1mean}, but with each $\bfv_j$ and
$\psi_{jj}$ being a nonlinear function of $x_j$.

\section*{Acknowledgements}
CW and AN would like to acknowledge the funding provided by the
UK Government's Defence \& Security Programme in support of the
Alan Turing Institute. 
The work of CW is supported in part by EPSRC grant
EP/N510129/1 to the Alan Turing Institute. CN is supported by 
a PhD studentship from the EPSRC CDT in Data Science  EP/L016427/1.
CW thanks Kevin Murphy for a useful email discussion.

\bibliographystyle{apalike}
\bibliography{fa_missing_refs}

\end{document}